\documentclass[runningheads]{llncs}
\usepackage{graphicx}
\usepackage{bbding}
\begin{document}
\title{A Semantic Indexing Structure for Image Retrieval}
%
%
\author{Ying Wang \and
Tingzhen Liu
\and
Zepeng Bu
\and
Yuhui Huang
\and
Lizhong Gao   
\and
Qiao~Wang {\Envelope}}
\authorrunning{Y. Wang et al.}
%
\institute{School of Information Science and Engineering, Southeast University, Nanjing, China 
\email{\{wying,lzgao,qiaowang\}@seu.edu.cn}}
\maketitle              
\begin{abstract}
In large-scale image retrieval, many indexing methods have been proposed to narrow down the searching scope of retrieval. The features extracted from images usually are of high dimensions or unfixed sizes due to the existence of key points. Most of existing index structures suffer from the dimension curse, the 

 unfixed feature size and/or the loss of semantic similarity. In this paper a new classification-based indexing structure, called Semantic Indexing Structure (SIS), is proposed, in which we utilize the semantic categories rather than clustering centers to create database partitions, such that the proposed index SIS can be combined with feature extractors without the restriction of dimensions.  Besides, it is observed that the size of each semantic partition is positively correlated with the semantic distribution of database. Along this way, we found that when the partition number is normalized to five, the proposed algorithm performed very well in all the tests. Compared with state-of-the-art models, SIS achieves outstanding performance.

\end{abstract}
\section{Introduction}

During the process of large-scale instance image retrieval, if the feature from the query image is compared with the features of database images in sequential way, it will consume not only a lot of computing resources but also plenty of time, which is inconvenient in industrial and commercial applications. In order to find the objects similar to the query image fast and ignore the unimportant ones, Approximate Nearest Neighbor Search (ANN Search) has attracted much attention. The basic idea of ANN Search is to partition or sort the database features in advance so as to narrow the searching scope. 

The main methods of ANN Search can be divided into three categories: tree-based indexing structure, hash-based indexing structure~\cite{simhash}~\cite{hash 1}~\cite{minhash}, and vector-quantization-based indexing structure~\cite{VQ}~\cite{survey PQ}~\cite{VLQ-ADC}~\cite{faiss}. The indexing structures based on tree, such as KD-tree and R-tree, are aimed to build a binary tree or poly tree by sorting the feature vectors in databases. The query vector can find the close leaf nodes fast by following the branches rather than traversing all leaf nodes. However, when data dimension is high (higher than 20 or so), no tree index structure can be significantly faster than a linear scan of the entire data set~\cite{high demension curse 4}, which is one of typical examples of dimensionality curse. For image retrieval, the produced feature vectors usually have very high dimensions, which leads to the limitation of tree structure.  

The indexing structures based on Hash, such as Simhash~\cite{simhash} and Minhash~\cite{minhash}, are designed with the goal of compressing the objects into binary code, so that the dimensionality curse can be convinced and the Hamming distance can be used to measure the similarity between different samples. The classical Hash algorithms lost the distribution of the original objects, which inspired the creation of Locality Sensitive Hashing(LSH)~\cite{LSH 2}. However, LSH is not very suitable for image processing~\cite{hash 2} since that the features are already approximation of images, if we further compress the features through Hash encoding to fit the indexing structure, it will cause a serious loss of both accuracy and semantic information. Besides, feature extractors based on key point detection don’t have definite size~\cite{sift}~\cite{delf}. The challenge about how to adapt the unfixed-size features to the Hash indexing structures still exists.

On the other hand, the indexing structures based on vector quantization, such as VLQ-ADC~\cite{VLQ-ADC} and Faiss~\cite{faiss}, use clustering or dimensionality reduction techniques to compress data and narrow searching range on vector scale. Compared with Hash encoding, vector operations can remain more information. Through the quantization, features based on key points which have uncertain size can be regularized into fixed-dimensional vector so as to be applied to tree indexing or Hash indexing. It should be stressed that  the cluster centers are not the same as semantic classification. The cluster centers are too abstract and do not map to the specific meaning in the real world. As a result, the vectors compressed often lose semantic information and the index built from them will suffer serious decline on accuracy. Further more, because the vector quantization methods are totally abstract, it becomes extremely difficult to optimize the structure without semantic supervision.
\begin{figure}[t]
\includegraphics[width=\textwidth]{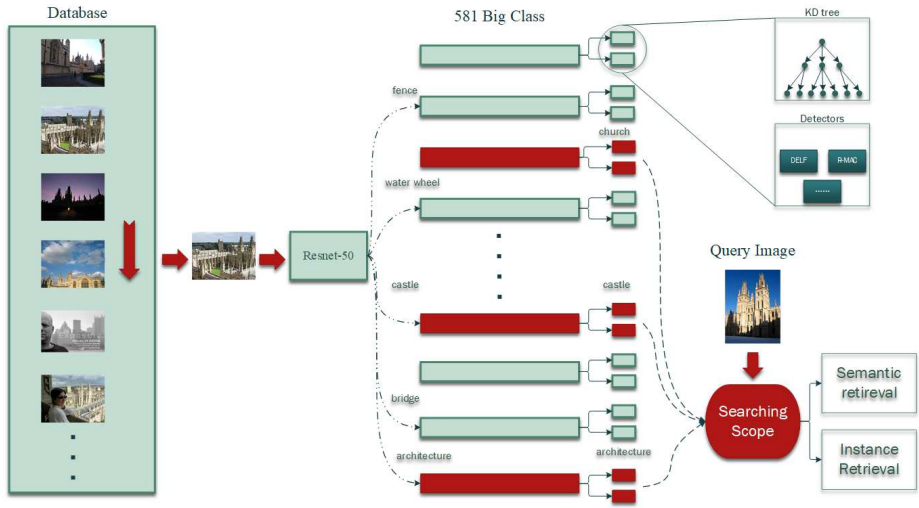}
\caption{We propose to use semantic classification instead of cluster centers to reduce the search range and proposed a new indexing structure named Semantic indexing structure(SIS). The system here is an example where $\alpha=5$ and $\beta=3$.} \label{fig8}
\end{figure}

Many image classification models~\cite{alexnet}~\cite{vggnet}~\cite{resnet} can get the semantic class labels from a picture. So we thought of using semantic categories to replace cluster centers in index structure and propose a new indexing structure named Semantic indexing structure(SIS). As we all know, the top-5 accuracy of image classification under the ILSVRC2012 data set is over 90\% with 1000 classes. So we propose the following two hypotheses. Firstly, at least one of the top-5 predictions of two similar images will be the same. As a result, if we throw the database images into semantic classes, We only need to select the five class partitions related to the query image as the searching scope, and lots of unimportant samples will be ignored. The minimal loss of ground-truth in our experiments proves this hypothesis. Secondly, even if the two close images do not belong to these 1000 classes semantically, there must be overlaps in the two groups of five partitions. Thus no matter what is the semantic class contribution of the data set, narrowing the search scope is effective. Oxford5k and Paris6k are two benchmark data sets for image retrieval. There are lots of building images, which is different from ILSVRC2012 for image classification with wider categories. Our experiments show the partitions based on 1000 classes still work on the image retrieval benchmark data sets. All in all, the results of our experiments have confirmed that our two hypotheses are basically true. And SIS is designed based on these two hypotheses.

The contributions of this paper can be summarized as three aspects:
\begin{enumerate}
\renewcommand{\labelenumi}{(\theenumi)}
    \item We analyze the shortcomings of current index construction methods, the dimensionality curse, the inapplicability of irregular features and the loss of semantic information. To solve these problems, we propose a new method to build index through semantic classes. For our approach, we put forward two hypotheses and use experiments to prove that they are basically true.
    \item We propose a new classification-based indexing structure called Semantic indexing structure (SIS), which can significantly narrow the searching scope and speed up image retrieval. Any ANN Search model will bring accuracy loss, but the loss of SIS is acceptable and much better than the state-of-the-art models.
    \item SIS has good versatility. It can be combined with any feature extractor, no matter the fixed-size feature, such as R-MAC~\cite{R-MAC}, or the key-point-based feature, such as SIFT~\cite{sift} and DELF~\cite{delf}. And our method can provide some inspiration on how to combine higher semantic features and lower number features.
\end{enumerate}

\section{Related Work}
\subsubsection{Image Classification} Plenty of models based on CNN~\cite{alexnet}~\cite{vggnet}~\cite{resnet}  show excellent performance for Image Classification since ILSVRC competition started in 2010. These models are mostly trained and tested on ILSVRC2012 data set .They can output the confidence vector of the image under 1000 class labels. Take the Resnet-50~\cite{resnet} as an example. The Top-1 accuracy is up to 79.26\% and the Top-5 accuracy is up to 94.75\%. We find that even if there is only one benchmark ground-truth, the other classes annotated as negative in top-5 usually also have semantic relationship with the predicted image. Therefore, we propose the method building multiple subspaces in the indexing structure based on the classes and record each image from databases to five subspaces of the index which are semantically related to the image content. During the stage of retrieval ,we choose the five subspaces related to the query image as the searching scope, which is undoubtedly smaller than the scope through a linear scan.
\subsubsection{Image Feature Extraction} Most of early feature extraction algorithms are mainly based on key point detection, such as SIFT~\cite{sift} and ORB~\cite{orb}. These algorithms use key points to locate the corners in the image and use descriptors to record the gradient details around the key points. The feature extracted by these methods pays too much attention on the detail texture information and loses the higher global contour information or semantic information. Further more, the size of feature vectors is unfixed as a result of the uncertain amount of the key points. This is unfriendly for the existing index structure. But our SIS index do not have this limitation. Any feature extractor can be applied in this structure and greatly accelerated. In recent years, CNN models~\cite{alexnet}~\cite{vggnet}~\cite{resnet} 
have made great progress in semantic information extraction, but the detail texture information are lost. Therefore, how to combine the higher semantic feature with the lower texture information becomes a challenge. The algorithms R-MAC~\cite{R-MAC} and DELF~\cite{delf} based on this idea perform very well in the application of image retrieval. Due to their calculations are too complicated, the efficiency is too low to apply them to the realistic retrieval. On the contrary, our index structure SIS will help a lot to alleviate this problem.

\section{Proposed Method}
\subsection{Pre-processing of classification labels}
The images are generally predicted under 1000 categories during classification. We find some of the categories is the subset of another major category. Expert experience is needed to distinguish them. For example, ordinary people cannot separate Indian elephants from Asian elephants. So we merge the 1000 categories into 581 major categories which are more common in daily life. In our scheme, we classify them into 581 Big Class. The partition of the data set and the construction of the index are both based on these 581 semantic tags.

\begin{figure}[t]
\includegraphics[width=\textwidth]{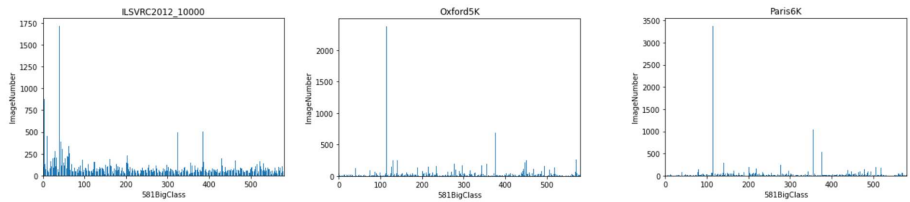}
\caption{The semantic distribution of data sets under 581 Big Classes. Each data set has its own focus. For example, the ILSVRC2012 data set has more animal pictures, while the distributions of Oxford5K and Paris6K are similar containing much more buildings.} \label{fig1}
\end{figure}

\subsection{Building the Index Structure of Database}
We use Resnet-50 models to predict each image's probability under the 1000 labels in the database. The probabilities under the 581 Big Class is gotten by adding. The calculation is written as:

\begin{equation}
\centering
P^b_i = \sum_{j=1}^{N^c_i}P^c_j
\end{equation}
Where $P^b_i$ represents the probability of the event that the predicted image belongs to the Big Class $i$. $N^c_i$ means the number of the children categories under the Big Class $i$. $P^c_j$ is the predicted value of the children class $j$ under the Big Class $i$.

For the images in the database, we can get the top $\alpha$ Big Class labels and the corresponding confidence. $\alpha$ presents the number of partitions one image will be recorded in the index structure. It usually equals 5 because the Top-5 accuracy of Resnet-50 is up to 94.75\%, which may also be 3 or 10 etc. For the query image, we can get the top $\beta$ Big Class labels and the relevant probability. $\beta$ presents the number of partitions we search. It also equals 5 in the common situation, which can be changed to 3 or 10 etc.
\begin{figure}[t]
\includegraphics[width=\textwidth]{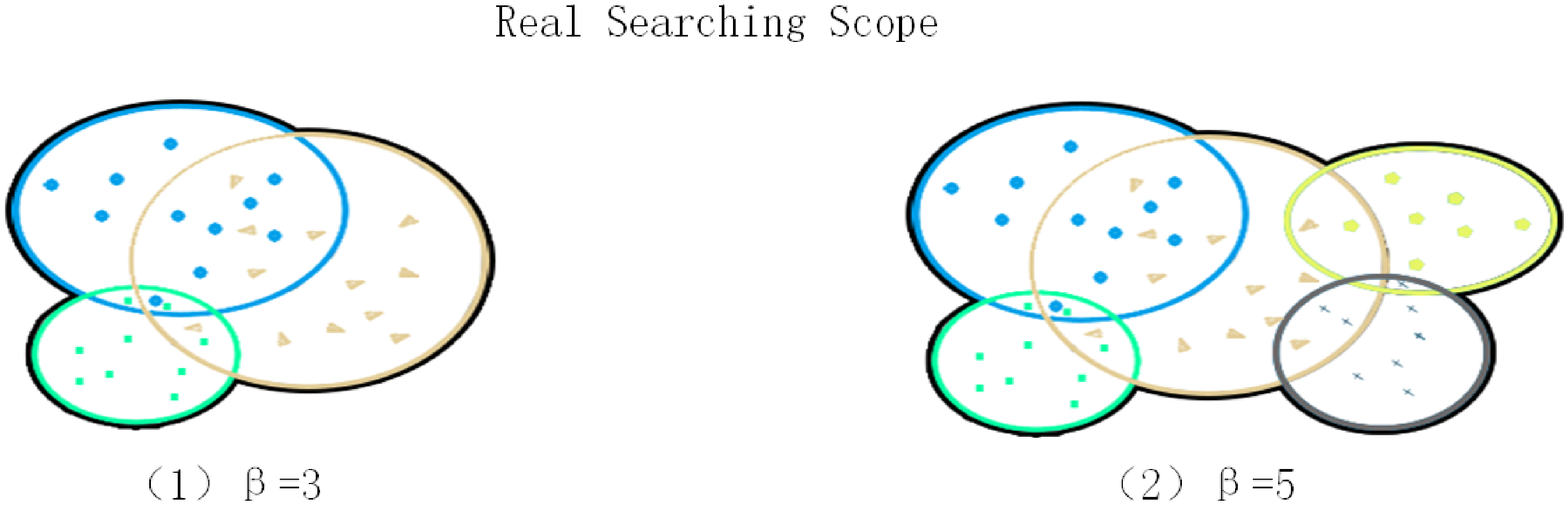}
\caption{The searching scope is the union of $\beta$ blocks. These blocks overlap each other. So the $\beta$ blocks have repeated elements. These elements need to be removed out during the process of retrieval to avoid the waste of time.} \label{fig2}
\end{figure}
The images in the database are recorded in SIS for $\alpha$ times and the total number of the nodes is $\alpha \cdot N_d$. The $\beta$ searching partitions are picked up from the total 581 blocks. These blocks overlap each other. So the $\beta$ blocks have repeated elements. These elements need to be removed during the process of retrieval to avoid the waste of time. The searching scope is the union of five blocks. If the amounts of images belong to each partition are almost the same, the expectation of the searching scope size is as blow:
\begin{equation}
\centering
E_s = \frac{\beta}{N_p} \cdot N_{d} \cdot \alpha-N_{repeat}= \frac{\beta}{N_p} \cdot N_{d}
\end{equation}
where $E_s$ is the expectation of the amount of images searched, which is smaller than the size of the database $N_{d}$ so as to speed up the retrieval. $N_p$ is the number of the blocks the index structure have. In SIS, $N_p$ equals 581. $N_{repeat}$ is the number of the repeated elements in the searching scope. But the situation that images distribute evenly under 581 Big Class is too ideal. The actual situation is that the image distribution of each data set has its own focus. For example, the ILSVRC2012 data set have more animal pictures, while the Oxford5K and Paris6K are similar containing much more buildings. (Database semantic distribution see Fig.~\ref{fig1}). As a result, the real searching scope size is usually bigger than $E_s$. We can use the ratio $R_{scope}$ between $S_{real}$ and $N_d$ to evaluate the acceleration effect of the index structure. At last, we may write the above measurements as follows,
\begin{equation}
S_{real}=\#(\cup_{i=1}^{\beta}Block_i),
\end{equation}

\begin{equation}
E_s<S_{real}<N_d,
\end{equation}

\begin{equation}
R_{scope}=\frac{S_{real}}{N_d}.
\end{equation}

\subsection{Semantic Retrieval}
At present stage,  the image retrieval can be performed under the narrowed  scope, since that the searching range is narrowed down through the classification-based index. Here, we build KD tree under each semantic block by using the images' confidence of belonging to the current session. Thus we can quickly find out the sample with the highest confidence under the partition generated by the KD tree. For the query image, we can provide $\beta$ images which are semantically related to the query one in this method.

\begin{figure}[t]
\includegraphics[width=\textwidth]{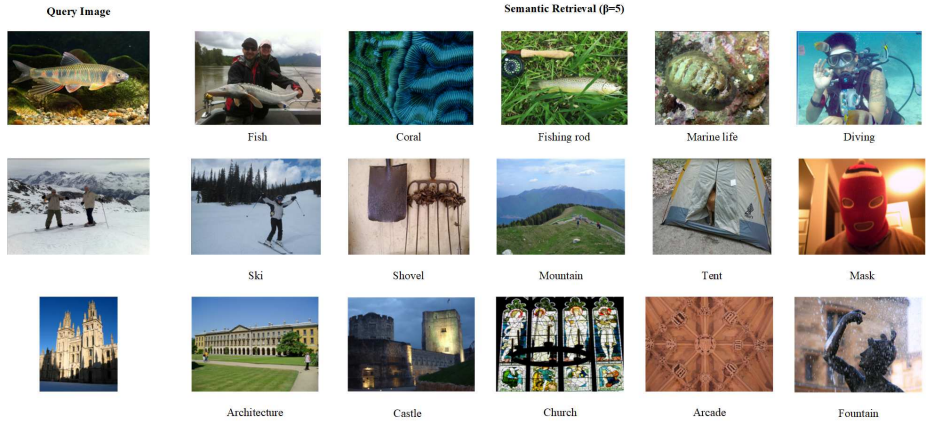}
\caption{Semantic Retrieval can display semantic information to imitate human associations and inferences. But the level of the semantic information is too high. The output images of the semantic retrieval are similar semantically to query image but not similar in details.} \label{fig3}
\end{figure}

As illustrated in Fig.~\ref{fig3}, the result of Semantic Retrieval is conducive to the understanding and association of images.

\subsection{Instance Retrieval}

Semantic Retrieval can provide semantic information to imitate human associations and inferences. But the level of the semantic information is too high. The output images of the semantic retrieval are similar semantically to query image but not similar in details. So we propose that the semantic partition should be combined with the feature extractors which can describe the texture details. We call the feature extraction algorithms as detectors. Because our index structure SIS has no limitation on the feature vectors' shape, any excellent detectors can be applied in SIS to obtain an increase in computational efficiency. Our structure can speed up the detectors significantly compared with linear scanning, while the accuracy will decline slightly at the same time. 

In instance image retrieval, the images in database are sorted by distance or similarity. The front images get higher scores if the images are ranked by similarity, while the front images get lower scores if the images are ranked by distance. The $\beta$ blocks selected as search scope also have their own confidence converted from the query image. We find some of the images from unimportant partitions will appear in the leading part of the retrieval list. So we use the probability information to adjust the score in order to alleviate the problem of decreased accuracy. However, the experimental results show this method is not acceptable. We think the cause of the problem lies in the operation of ignoring repeated elements and the appropriate adjustment on the score. We still illustrate it here in this work. 

In what follows we denote $ S_{C}=S\times C$, if $Score=Similarity$ but $S_C=S/C$, if $Score=Distance$, where $S$ is the score used to sort the images, $C$  the confidence of the current block, and  $S_C$  the adjusted score by the confidence value.

\begin{figure}[t]
\includegraphics[width=\textwidth]{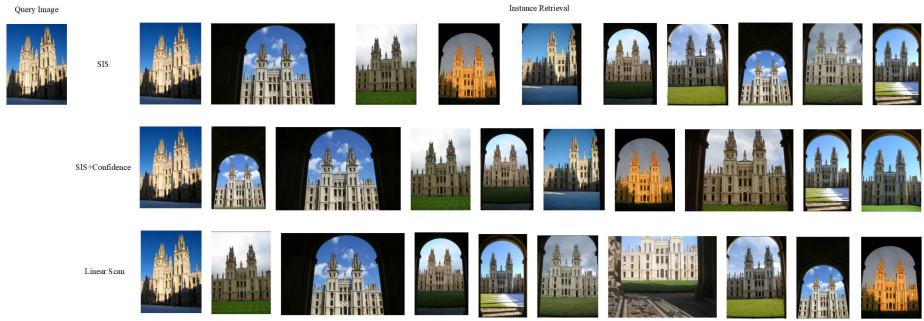}
\caption{The retrieval outputs from linear scanning, SIS and SIS with Confidence are shown here. The Data set is Oxford5K, and $\alpha=5$, $\beta=5$} \label{fig4}
\end{figure}

\section{Experiments}
\subsection{Evaluation}
We use $mAP$ and $Recall$ to evaluate the accuracy of instance image retrieval. The real searching scope can reflect the acceleration efficiency of the index structures. So we set $Ratio_{scopt}$ to stands for the mean real searching scope ratios,  $Ratio_{time}$  the mean ratio of the query time in index and the time in linear scanning. The evaluation systems are list as below,
\begin{equation}
precision_k^j=\frac{\sum_{i=1}^{k}R^j_e(i)}{k},
\end{equation}
\begin{equation}
AP_j=\frac{1}{N_{gt}^j}\sum_{k=1}^{N_{rs}^j}precision^j_k,
\end{equation}
\begin{equation}
mAP=\frac{1}{N_t}\sum_{j=1}^{N_t}AP_j,
\end{equation}
where $R^j_e(i)$ means the ground truth relevance between the $j$-th query image in the test set and the $i$-th ranked image in the result list. For the range of $R^j_e(i) \in [0,1]$, $0$ is irrelevant, and $1$ highly related. $N_t$ is the size of the test set, and $N^j_{gt}$  the amount of the ground truths of $j$-th query image. Finally, $N^j_{rs}$ is the real size of the searching scope of $j$-th query image in the test set, and $mAP$ the mean value of all query image's $APs$. We denote
\begin{equation}
Recall=\frac{1}{N_t}\sum_{j=1}^{N_t}\frac{N^j_{rc}}{N^j_{gt}},
\end{equation}
where $Recall$ can reflect the current index structure's ground truth loss rate so as to evaluate the rationality of the narrowed scope, $N^j_{rc}$  the amount of the recalled ground truths when we search the $j$-th query image in the test set. Then we define
\begin{equation}
Ratio_{scopt}=\frac{1}{N_t}\sum_{j=1}^{N_t}\frac{S^j_{real}}{N_d},
\end{equation}
in which $Ratio_{scopt}$ means the scope narrowing rate of the index structure, ${S^j_{real}}$  the size of the real searching scope when we search the $j$-th query image, and $N_d$  the size of the database. We define 
\begin{equation}
Ratio_{time}=\frac{1}{N_t}\sum_{j=1}^{N_t}\frac{t^j_{idx}}{t^j_{whl}},
\end{equation}
where $Ratio_{time}$ measures the mean ratio between the query time with index structure and the query time in sequential way,  ${t^j_{idx}}$  the query time with index structure when we search the $j$-th query image,  $t^j_{whl}$ the query time in sequential way when we search the $j$-th query image.

\subsection{Dynamic Query Expansion of SIS}

\begin{figure}[h]
\includegraphics[width=\textwidth]{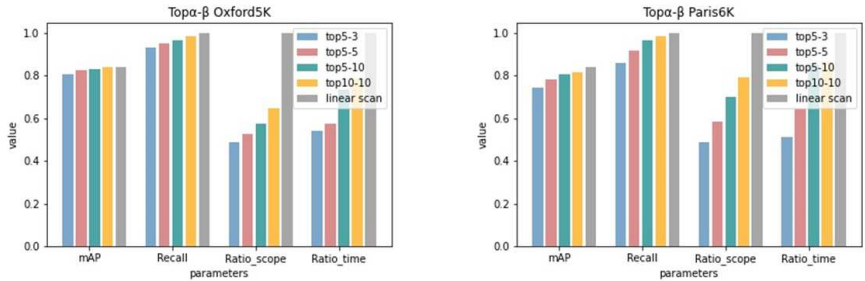}
\caption{The higher DEF $\alpha$ or QEF $\beta$ is ,the searching scope is bigger. Accompanying this expansion is the increase in accuracy and the decrease in acceleration efficiency.} \label{fig5}
\end{figure}

As mentioned above,  two parameters $\alpha$ and $\beta$
 can be conditioned to adjust the searching scope of the image retrieval. We call $\alpha$ the Database Expansion Factor (DEF), and $\beta$ the Query Expansion Factor (QEF), respectively. The higher DEF or QEF corresponds to the bigger searching scope. For the trade-off between accuracy and acceleration efficiency, we point that the dynamic balance between accuracy and speed can be obtained by suitable setting parameters (See Fig.~\ref{fig5}). We choose $\alpha=5$ and $\beta=5$ to carry out follow-up experiments.

It should be noticed is that SIS can automatically adjust searching scope according to DEF and QEF. The adjustment  is driven by the data in databases and query image, which is an advantage of SIS, while the search ratio of other index has to be set artificially relying on subjective experience.
\subsection{Comparison with State-of-Art Methods}
The $mAP$ of DELF~\cite{delf} is up to 0.848 on Oxford5K. Thus,  we begin with  DELF as the detector of SIS  with $\alpha=5$ and $\beta=5$. Here the searching scope is narrowed to 52.5\%, and the speed will be nearly doubled. Under this index scheme, the accuracy decline will inevitably occur from 0.848 to 0.826. If we expand the scope by setting $\alpha$ as 10 and $\beta$ as 10, the $mAP$ value will raise to 0.841 , while the scope expanded to 64.8\%(See Fig.~\ref{fig5}). 
\begin{table}[h]
\caption{Comparison with State-of-Art Methods when $\alpha=5$ and $\beta=5$ in Oxford5K}\label{tab2}
\begin{tabular}{|l|l|l|l|l|l|l|l|}
\hline
Index & Detectors &$b_{total}$&$b_{selected}$& $mAP$ & $Recall$  & $Ratio_{scope}$ & $Ratio_{time}$ \\
\hline
Linear Scan &  DELF+Voc Tree~\cite{voc} &-&-& 0.561 & 1  & 1 & 1\\
Faiss-IVFPQ~\cite{faiss} &  DELF+Voc Tree~\cite{voc} &100&10& 0.554
 & 0.805  & 0.126 & 0.368\\
Faiss-IVFFLAT~\cite{faiss} &  DELF+Voc Tree~\cite{voc} &100&10& 0.556 & 0.737  & 0.089 & 0.368\\
Faiss-IVFPQ~\cite{faiss} &  DELF+Voc Tree~\cite{voc} &100&52& 0.529
 & 0.997  & 0.645 & 0.333\\
Faiss-IVFFLAT~\cite{faiss} &  DELF+Voc Tree~\cite{voc} &100&52& \bfseries0.596 & 0.983  & \bfseries0.476 & 0.325\\
SIS+Confidence &  DELF+Voc Tree~\cite{voc} &581&5& 0.524 & 0.953 & 0.525 & 0.441\\
SIS &  DELF+Voc Tree~\cite{voc} &581&5& 0.582 & 0.953& 0.525
& 0.451\\
\hline
Linear Scan &  R-MAC~\cite{R-MAC} &-&-& 0.551 & 1  & 1 & 1\\
Faiss-IVFPQ~\cite{faiss} &  R-MAC~\cite{R-MAC} &100&52& 0.436 & 0.990  & 0.595 & 0.052\\
Faiss-IVFFLAT~\cite{faiss} &  R-MAC~\cite{R-MAC} &100&52& \bfseries0.551 & 0.990  & 0.595 & 0.053\\
SIS+Confidence &  R-MAC~\cite{R-MAC} &581&5& 0.516 & 0.953 & 0.525 & 0.268\\
SIS &  R-MAC~\cite{R-MAC} &581&5& 0.546 & 0.953& \bfseries0.525
& 0.397\\
\hline

Linear Scan &  BoDVW~\cite{bow}&-&-& 0.776 & 1  & 1 & 1\\
Linear Scan &  ILRGAN~\cite{ILRGAN}&-&-& 0.793 & 1  & 1 & 1\\
Linear Scan &  DELF ~\cite{delf} &-&-& 0.848 & 1  & 1 & 1\\
Faiss &  DELF~\cite{delf} &-&-& unable& unable& unable& unable\\

SIS+Confidence &  DELF~\cite{delf} &581&5& 0.799 & 0.953& 0.525
& 0.586\\
SIS &  DELF~\cite{delf} &581&5& \bfseries0.826 & 0.953 & \bfseries0.525 & 0.574\\
\hline
\end{tabular}
\end{table}

When we apply DELF to Faiss~\cite{faiss} for performance comparisons, a problem arises. Only features with fixed size can be applied to Faiss. The DELF feature vector's size relies on the number of key points. So we choose Vocabulary tree~\cite{voc} to perform vector quantization on DELF. The feature size becomes 341. We padded zeros at the end of the vector to expand the size to 512 so as to fit Faiss. The Ransac similarity evaluation DELF provided can't fit Faiss, either. So the $L_2$ distance relpaces it, while only $L_2$ distance and cosine similarity can be applied in Faiss. If we set the total amount of the blocks to 100 and the number of the searched blocks to 10 as the official demo did, a heavy loss of $Recall$ will happen. Only 80.5\% and 73.7\% of the ground truths are remained in Faiss-IVFFLAT and Faiss-IVFPQ. We inferred from the $Ratio_{scope}$ 0.525 of SIS with $\alpha=5$ and $\beta=5$ that the selected block number $b_{selected}$ should better be 52. The results confirm the performance. The accuracy of IVFPQ is lower than SIS and that of IVFFLAT is higher than SIS. 58.2\% and 59.6\% are very close and both far lower than the accuracy in SIS with DELF as a result of vector quantization shortcomings. The closeness proves that the narrowed scope of SIS is close to the ranked partitions divided by clustering in abstract vector space, thus our index structure is reasonable. And there is still some optimizing space for each semantic block in SIS. 

One thing should be noticed is that $b_{selected}$ in SIS can always be 5 no matter what is the dataset semantic distribution, but the number 52 in Faiss have to be changed if the dataset or the query image changes. Because the samples distribute  evenly across all partitions in traditional indexes. There is no way to find the best $b_{selected}$ without the help of SIS in the real application activities. We call that the adaptability driven by data of SIS. This is the advantage of our structure. R-MAC feature has the fixed dimension of 512, and can be compared by cosine similarity. The experiment on R-MAC get similar results to DELF+Vocabulary Tree, so we don't analyze them here.

All in all, SIS has three advantages. First, any feature detector and distance function can fit SIS. Second, $b_{selected}$ can remain 5 or 10 in SIS, which can adapt the distribution of database by itself. Third, user can choose between the accuracy and the scope narrowing rate by $\alpha$ and $\beta$. In theory, $Ratio_{scope}$ will be smaller if the distribution of database on semantic classes is 
wider and evener. Large-scale experiments will be carried out in future work. The result of $Recall$ and $mAP$ proves that the hypotheses mentioned in Introduction part are reasonable.

\section{Conclusions}

In this paper, we propose an index structure named SIS. The SIS provide a new way to combine high-level semantic information and low-level texture information. SIS solves some problems other ANN searching methods bring, which result from the dimension limitation of other Index. SIS have excellent adaptability and versatility, which can be combined with any feature extractors and can adjust the searching scope by itself according to the distribution of image databases. The retrieval speed of SIS almost doubles, with only slightly decrease of accuracy. Further more, the decreased accuracy is still competitive compared with state-of-the-art methods.
%
%
%

\end{document}